\pdfoutput=1

\documentclass[11pt]{article}
\usepackage{amssymb}
\usepackage[table,dvipsnames]{xcolor}
\usepackage{wrapfig}
\usepackage{epigraph}

\newcommand{\blueshaded}[1]{\colorbox[HTML]{dfe5f2}{#1}}
\newcommand{\redshaded}[1]{\colorbox[HTML]{f9ede5}{#1}}
\newcommand{\greenshaded}[1]{\colorbox[HTML]{edf4e6}{#1}}
\newcommand{\yellowshaded}[1]{\colorbox[HTML]{fff8e5}{#1}}

\usepackage[utf8]{inputenc}  
\usepackage[T1]{fontenc}     

\usepackage[final]{acl}

\usepackage{times}
\usepackage{latexsym}

\usepackage{xspace}
\newcommand{\AutoUrbanCI}{\textbf{\texttt{UrbanCIA}}\xspace} 
\newcommand{\AgentName}[1]{\texttt{#1}}

\newcommand{\shline}{\noalign{\global\arrayrulewidth=1.5pt}\hline}
\usepackage{pifont}
\usepackage{enumitem}
\usepackage{amsmath}
\usepackage{array,float}
\usepackage{listings}
\usepackage{arydshln}
\usepackage[T1]{fontenc}

\usepackage{microtype}

\usepackage{inconsolata}

\usepackage{graphicx}

\usepackage{etoolbox}
\usepackage{changepage} 

\usepackage{tabularx}
\usepackage{multirow}

\newcolumntype{M}[1]{>{\centering\arraybackslash}m{#1}}
\newcolumntype{L}[1]{>{\raggedright\arraybackslash}m{#1}}
\usepackage{booktabs}

\newcommand{\lionlogo}{\raisebox{3pt}{\includegraphics[scale=0.014]{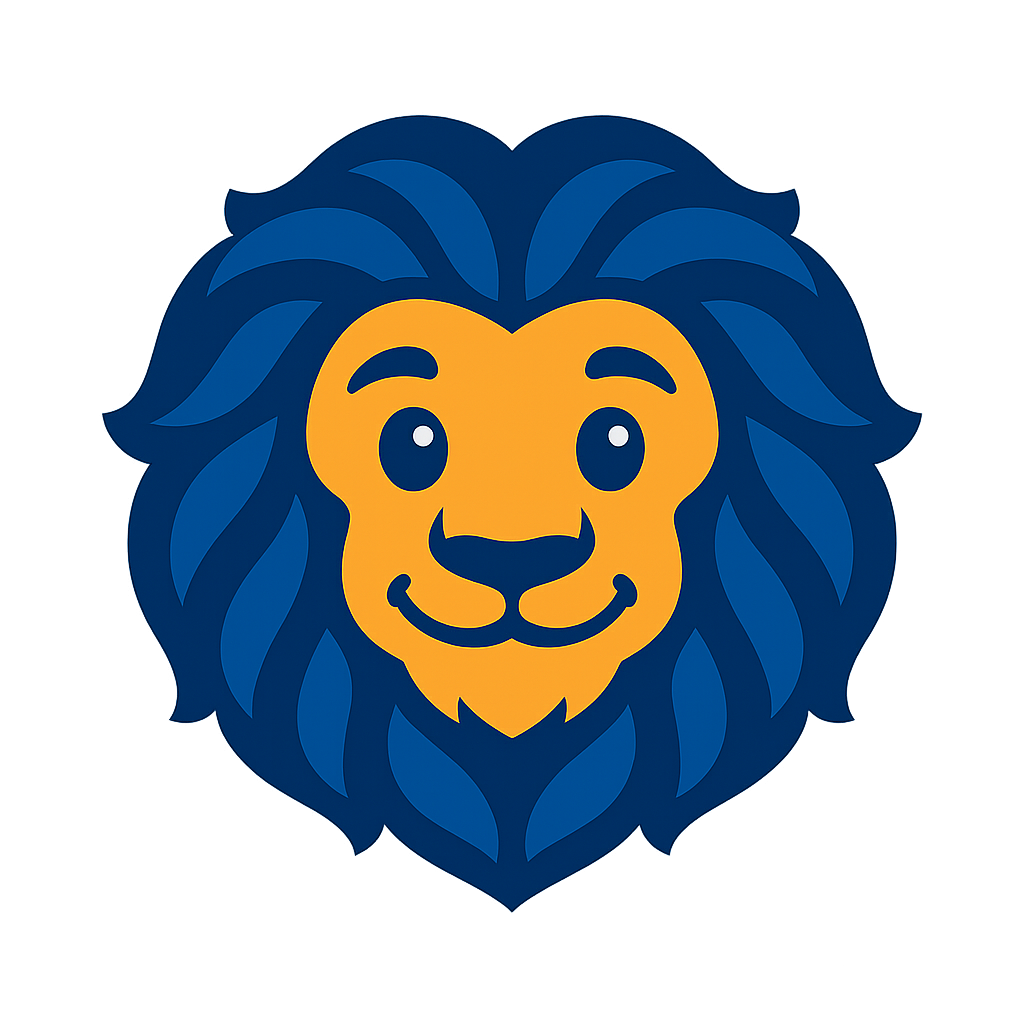}}}
\newcommand{\beaverlogo}{\raisebox{3pt}{\includegraphics[scale=0.015]{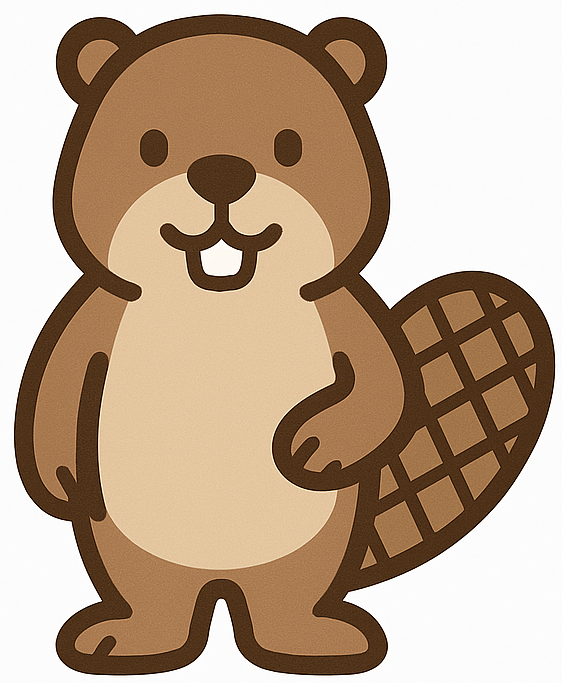}}}
\newcommand{\bridlogo}{\raisebox{3pt}{\includegraphics[scale=0.007]{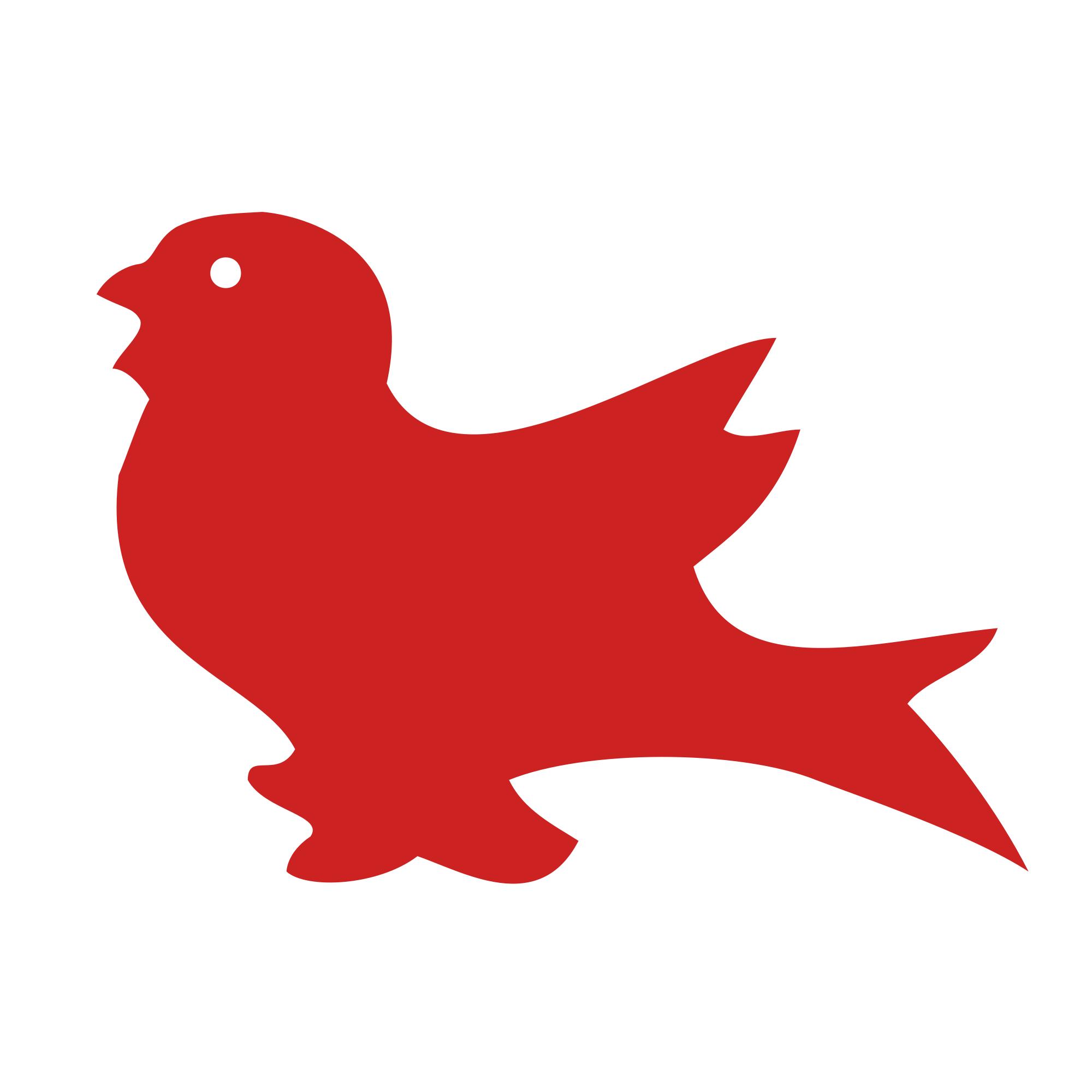}}}
\newcommand{\torchlogo}{\raisebox{3pt}{\includegraphics[scale=0.016]{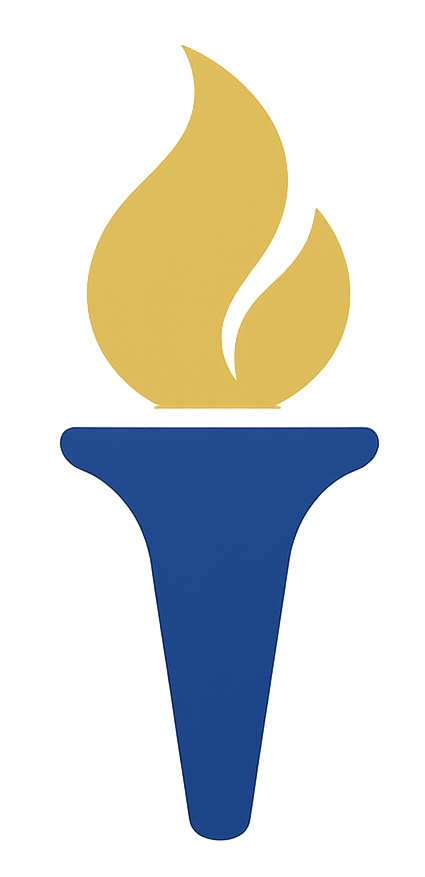}}}
\newcommand{\croclogo}{\raisebox{3pt}{\includegraphics[scale=0.013]{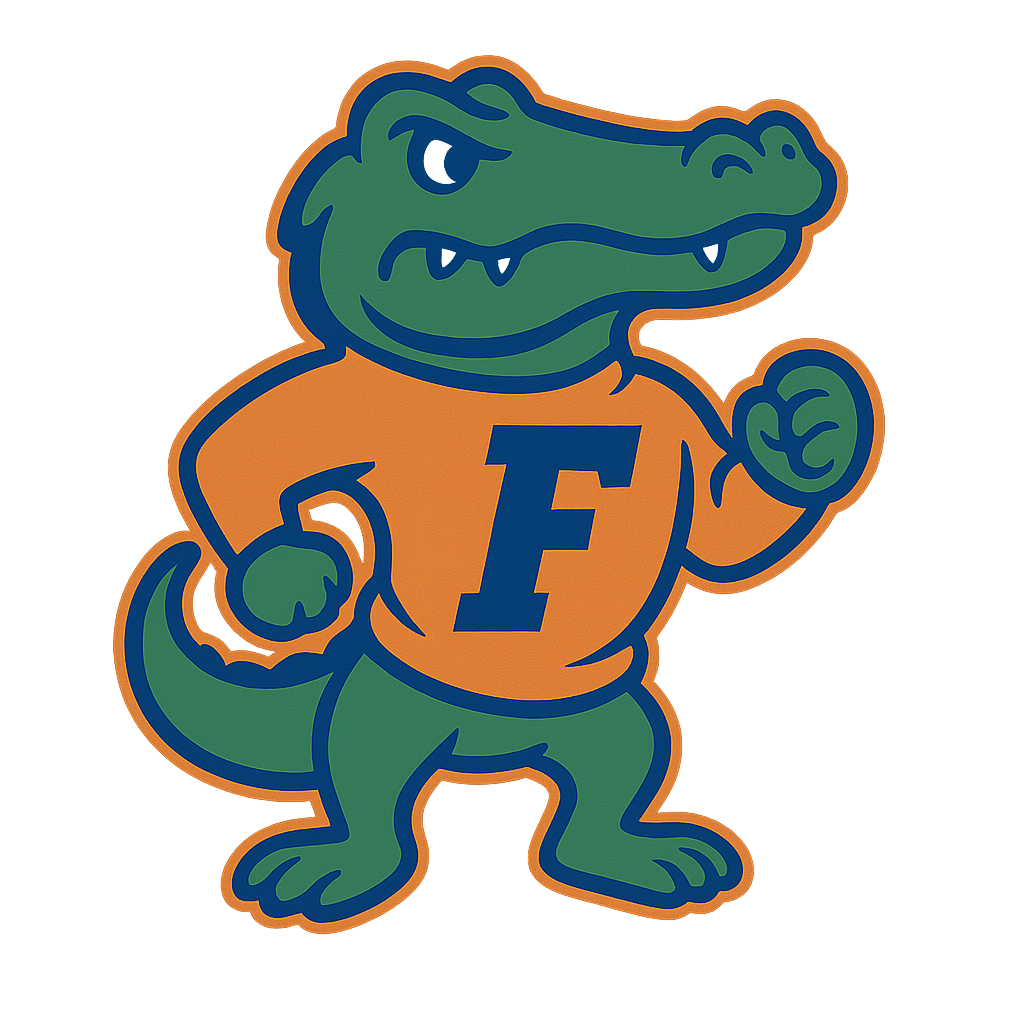}}}
\newcommand{\nus}{\lionlogo}
\newcommand{\mitlogo}{\beaverlogo}
\newcommand{\mcgilllogo}{\bridlogo}
\newcommand{\hkustlogo}{\torchlogo}
\newcommand{\uflogo}{\croclogo}
\newcommand\coauth{$^\diamondsuit$}

\definecolor{starmarkcolor}{HTML}{fda463}

\title{\raisebox{-1em}{\includegraphics[width=0.1\textwidth]{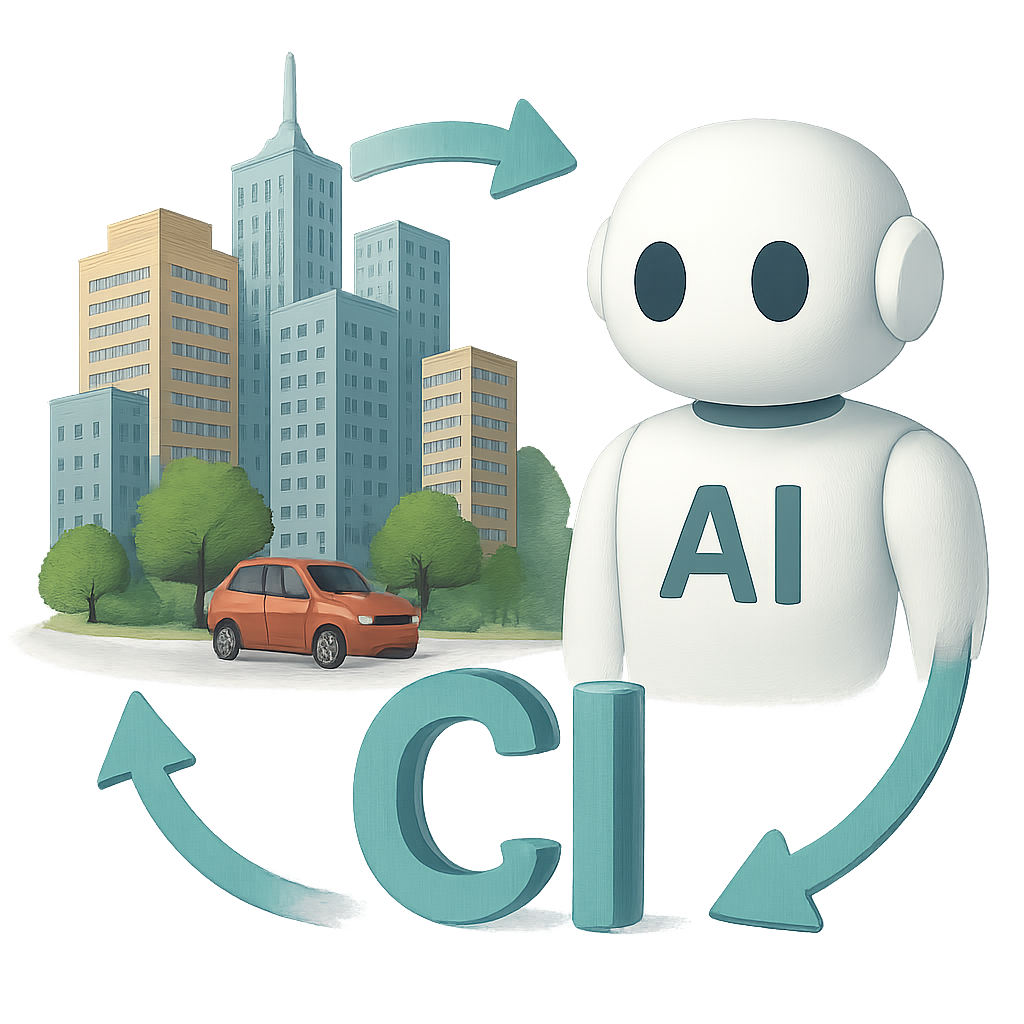}}
\LARGE Reimagining Urban Science: Scaling Causal Inference with Large Language Models
}

\author{\\ 
Yutong Xia\coauth{}\mitlogo{}\nus{} \quad 
Ao Qu\coauth{}\mitlogo{}  \quad 
Yunhan Zheng\mitlogo{} \quad 
Yihong Tang\mcgilllogo{} \quad 
Dingyi Zhuang\mitlogo{} \quad \\
Yuxuan Liang\hkustlogo{} \quad
Shenhao Wang\uflogo{} \quad
Cathy Wu\mitlogo{} \quad
Lijun Sun\mcgilllogo \quad
Roger Zimmermann\nus{} \quad 
Jinhua Zhao\mitlogo{} \quad\\
\mitlogo{}MIT \quad 
\nus{}NUS \quad 
\mcgilllogo{}McGill \quad 
\hkustlogo{}HKUST (GZ) \quad 
\uflogo{}UFL \quad 
\\\vspace{-0.1cm}
\small{
\texttt{\href{mailto://yutongx@mit.edu}{\{yutongx,qua,yunhan,dingyi,cathywu,jinhua\}@mit.edu}} \quad 
\texttt{\href{mailto://yihong.tang@mail.mcgill.ca}{yihong.tang@mail.mcgill.ca}}} \\
\small{
\texttt{\href{mailto://lijun.sun@mcgill.ca}{lijun.sun@mcgill.ca}}\quad 
\texttt{\href{mailto://yuxliang@outlook.com}{yuxliang@outlook.com}}\quad 
\texttt{\href{mailto://shenhaowang@ufl.edu}{shenhaowang@ufl.edu}}\quad 
\texttt{\href{mailto://dcsrz@nus.edu.sg}{dcsrz@nus.edu.sg}} 
}
}

\newcommand\blfootnote[1]{%
  \begingroup
  \renewcommand\thefootnote{}\footnote{#1}%
  \addtocounter{footnote}{-1}%
  \endgroup
}

\begin{document}

\maketitle

\vspace{1.5cm}
\begin{abstract}

Urban causal research is essential for understanding the complex, dynamic processes that shape cities and for informing evidence-based policies. However, current practices are often constrained by inefficient and biased hypothesis formulation, challenges in integrating multimodal data, and fragile experimental methodologies.
Imagine a system that automatically estimates the causal impact of congestion pricing on commute times by income group or measures how new green spaces affect asthma rates across neighborhoods using satellite imagery and health reports, and then generates comprehensive, policy-ready outputs, including causal estimates, subgroup analyses, and actionable recommendations.
In this Perspective, we propose \AutoUrbanCI, an LLM-driven conceptual framework composed of four distinct modular agents responsible for hypothesis generation, data engineering, experiment design and execution, and results interpretation with policy insights. We begin by examining the current landscape of urban causal research through a structured taxonomy of research topics, data sources, and methodological approaches, revealing systemic limitations across the workflow.
Next, we introduce the design principles and technological roadmap for the four modules in the proposed framework. We also propose evaluation criteria to assess the rigor and transparency of these AI-augmented processes. Finally, we reflect on the broader implications for human–AI collaboration, equity, and accountability.
We call for a new research agenda that embraces LLM-driven tools as catalysts for more scalable, reproducible, and inclusive urban research.
\blfootnote{\coauth Equal contribution.}  
\end{abstract}
\vspace{0.4cm}

\epigraph{
\normalsize``\textit{Cities have the capability of providing something for everybody, only because, and only when, they are created by everybody.}''}{
\normalsize\textit{Jane Jacobs}
}
\section{Introduction}

Cities are complex systems shaped by interconnected forces, from transportation congestion and housing affordability to climate resilience and equitable public services~\cite{owid-urbanization,un-sdg11,xia2022impact}. Scientific progress in urban research equips policymakers, planners, and communities with the knowledge to design evidence-based interventions that shape urban futures~\cite{batty1997computable}. 
At the heart of this scientific effort is the pursuit of causal understanding: uncovering how policies, infrastructures, and behaviors interact to produce measurable impacts~\cite{baum2015causal}.
Urban causal research typically follows a structured pipeline: beginning with a hypothesis, integrating diverse data sources, selecting appropriate methods, and interpreting results to inform decisions. For example, in assessing New York City’s congestion pricing policy~\cite{bbc2024congestion}, scientists may hypothesize that its impact on commute times differs by neighborhood income levels. This triggers a series of steps: collecting multimodal data such as subway logs, taxi trips, social media, and census records; applying methods like difference-in-differences with fixed effects~\cite{hernan2020causal}. The final analysis reveals not only whether commute times decreased, but whether the policy disproportionately benefits certain socioeconomic groups.
Such approaches have also enabled studies on the effects of high-speed rail~\cite{heuermann2019effect}, electric vehicle infrastructure~\cite{zheng2024effects}, and vaccine mandates~\cite{karaivanov2022covid}. 

However, this process remains labor-intensive, fragile, and unevenly accessible. We identify three major challenges:
1) \textit{Hypothesis discovery is slow and biased}: research questions are often manually crafted and shaped by disciplinary precedent or funding priorities, overlooking emerging issues and underrepresenting smaller cities or informal settlements~\cite{bell2009small,muller2025toward}; 2) \textit{Data complexity impedes integration}: although urban environments generate rich and diverse data streams (e.g., satellite imagery, mobility logs, text reports), integrating them into causal studies remains technically demanding, leading many studies to rely primarily on structured tabular datasets~\cite{raghavan2020data,hong2023reconciling}; 3) \textit{Experimentation is expensive and risk-prone}: causal inference depends on stringent assumptions, such as valid counterfactuals or exogenous variation, and many studies are abandoned when these conditions fail, discouraging exploration of high-risk, high-reward policy questions~\cite{baum2015causal,matthay2022causal}.
Taken together, these challenges underscore the need for more systematic, data-aware, and technology-enabled approaches to causal inference in urban contexts.
The structural limitations of current urban causal research raise a timely question: \emph{Can artificial intelligence help expand the scope, efficiency, and inclusivity of urban scientific discovery?}

Recent advances in Large Language Models (LLMs)~\cite{achiam2023gpt,touvron2023llama,team2023gemini,guo2025deepseek} and Multimodal Large Language Models (MLLMs)~\cite{wu2023multimodal} offer new possibilities for reimagining how urban causal research is conducted. LLMs exhibit strong capabilities in language understanding~\cite{radford2018improving}, reasoning~\cite{wei2022chain,jaech2024openai}, planning~\cite{huang2022language}, code execution~\cite{chen2021evaluating}, and tool use~\cite{schick2023toolformer,qintoolllm}. 
MLLMs extend these functions across diverse data modalities such as imagery~\cite{liu2023visual}, audio~\cite{chu2023qwen}, graph~\cite{tang2024graphgpt}, structured charts~\cite{han2023chartllama}, and unstructured reports~\cite{li2024extracting}, lowering the barriers to engaging with complex urban data at scale~\cite{liang2025foundation}. 
Despite these technical advances, their application to urban causal research remains fragmented, often limited to isolated tasks such as data cleaning, code generation, or report drafting.
Recent developments in agent-based LLM systems~\cite{lu2024ai,schmidgall2025agent,gottweis2025aicoscientist} suggest a promising direction: not using LLMs as end-to-end black boxes, but as specialized components within a coordinated reasoning system. 

\begin{figure*}[t]
    \centering
    \includegraphics[width=0.8\textwidth]{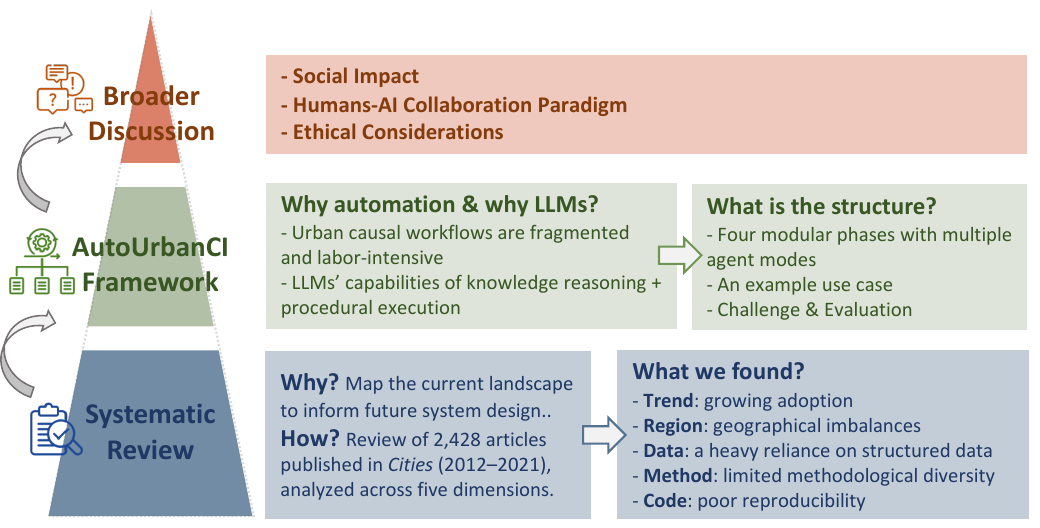}
    \caption{\textbf{Overview of the paper’s structure and logic.} A layered structure from review to framework and broader discussion. Arrows indicate how insights from one stage inform the next.}
    \label{fig:overview}
    \vspace{-1em}
\end{figure*}

In this Perspective, we argue that 
\textbf{LLMs offer a path to reimagine how we ask questions, extract evidence, and generate causal knowledge in cities - at scale, and with broader reach}.
To explore this potential, we introduce \AutoUrbanCI, a flexible and modular framework for \textbf{\underline{Urban}} \textbf{\underline{C}}ausal \textbf{\underline{I}}nference with \textbf{\underline{A}}gents. 
\AutoUrbanCI is a conceptual framework that supports the full urban causal pipeline: surfacing overlooked questions, assembling multimodal data, evaluating methodological assumptions, and generating interpretable outputs.
Compared to conventional manual workflows and isolated AI-assisted tools, \AutoUrbanCI offers a structured, extensible, and interpretable framework that supports the full urban causal inference process while preserving space for human judgment, domain expertise, and ethical deliberation.

The outline of this paper are shown in Figure~\ref{fig:overview}. Specifically, Section~\ref{sec:review} presents a systematic review of urban causal research, identifying key gaps in data, methods, and geographic coverage. Section~\ref{sec:framework} introduces \AutoUrbanCI, a modular agent-based framework that supports all stages of causal inference, with a concrete multi-agent system use case. We also discuss potential challenges and propose evaluation methods. Section~\ref{sec:discussion} reflects on how \AutoUrbanCI enables broader participation from civic groups, governments, and urban communities, and offer a perspective on the human-AI collaboration paradigm. Finally, Section~\ref{sec:sum} summarizes the paper and outlines future directions.
Our contributions are summarized as follows:
\begin{itemize}[itemsep=0pt, topsep=2pt]
\item  \textbf{A Comprehensive Systematic Review.} We conduct a large-scale empirical analysis of causal inference studies in urban research, covering over 2,400 articles across a decade from a flagship journal (i.e., \textit{Cities}), identifying key trends and gaps in the use of causal inference under the urban context.
\item  \textbf{A Modular Framework.} We introduce \AutoUrbanCI, a flexible, AI-augmented pipeline that supports all stages of causal analysis through modular agents spanning hypothesis generation, data preparation, experimental execution, and report writing, demonstrated through a concrete multi-agent system use case.
\item \textbf{A Multi-Dimensional Evaluation Protocol.} We examine methodological, ethical, and deployment challenges, and propose metrics for assessing the rigor, novelty, and generalizability of AI-generated causal research. 
\item \textbf{Towards Inclusive Urban Science.} We show how \AutoUrbanCI reduces the entry barriers to rigorous urban causal research, allowing broader participation by civic organizations, municipal governments, and community groups - advancing the vision of a city ``\textit{created by everybody}''~\cite{jacobs1961death}.
\end{itemize}

\section{Landscape of Causal Inference in Urban Studies}\label{sec:review}
This section provides a quantitative overview of the field based on a comprehensive empirical analysis. It identifies key trends in causal inference research across urban topics, data usage, methodological approaches, and geographic focus. These findings lay the foundation for designing \AutoUrbanCI, our proposed AI framework for automated causal research, ensuring it is aligned with actual research needs, practices, and gaps (see Appendix~\ref{app:review-method} for methodological details).

\subsection{Why a Systematic Review?}

\textbf{\textit{- Mapping the current landscape to inform future design.}}
To develop an effective AI system for causal inference in urban research, it is critical to first understand how such research is currently conducted. Urban systems are inherently complex, characterized by dynamic, interlinked processes spanning infrastructure, policy, society, and environment. Identifying causal relationships within such systems is essential for evaluating the impact of interventions, but the methodological choices, data modalities, and research foci within this domain remain unevenly distributed and often under-documented.

\begin{wrapfigure}{R}{0.5\textwidth}
  \includegraphics[width=\linewidth]{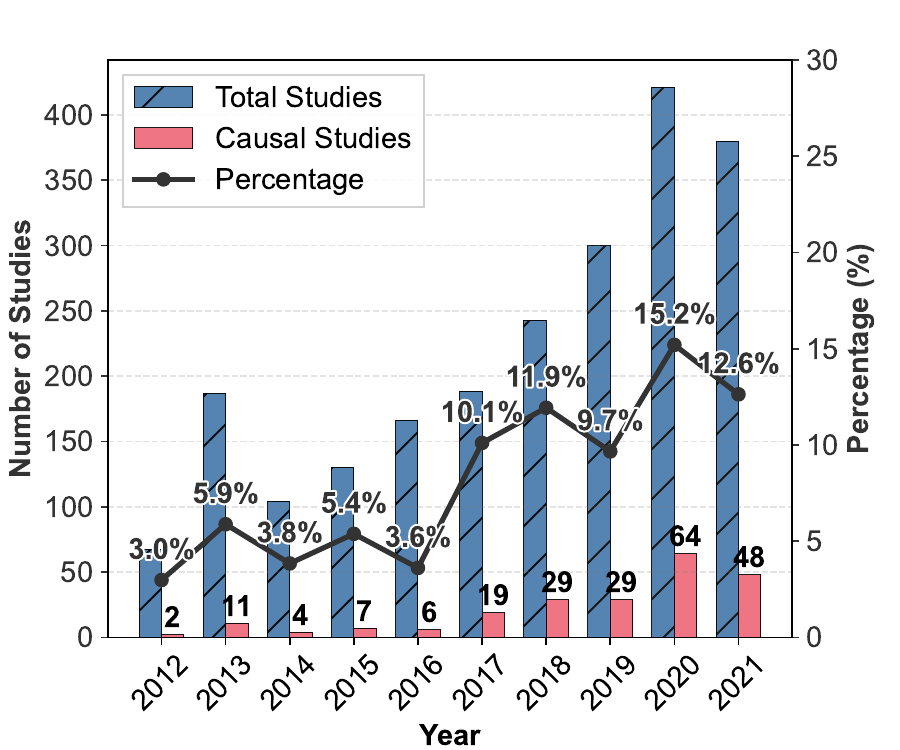}
  \vspace{-0.5cm}
    \caption{\textbf{Rising adoption of causal inference in urban research.} Trends in publication volume (blue bars) and causal inference usage (red bars) in \textit{Cities} from 2012 to 2021. The dotted line marks the growing share of studies employing causal methods over time.}
  \label{fig:trend}
    \vspace{-0.5cm}
\end{wrapfigure}

\subsection{How We Reviewed It?}
\textbf{\textit{- Data Collection.}} We conducted a systematic review of 2,428 peer-reviewed articles published in \textit{Cities}, a leading journal in urban research, spanning a 10-year period from 2012 to 2021. The related publication trend of urban causal inference studies is shown in Figure~\ref{fig:trend}.
Using a pipeline combining state-of-the-art LLMs, i.e., GPT-4o-mini, GPT-4o, and GPT-4.5~\cite{jaech2024openai}, and classical machine learning techniques, followed by expert human validation, we identified 219 articles that employed formal causal inference methodologies in their urban studies.

\begin{figure*}[t]
    \centering
    \includegraphics[width=1\linewidth]{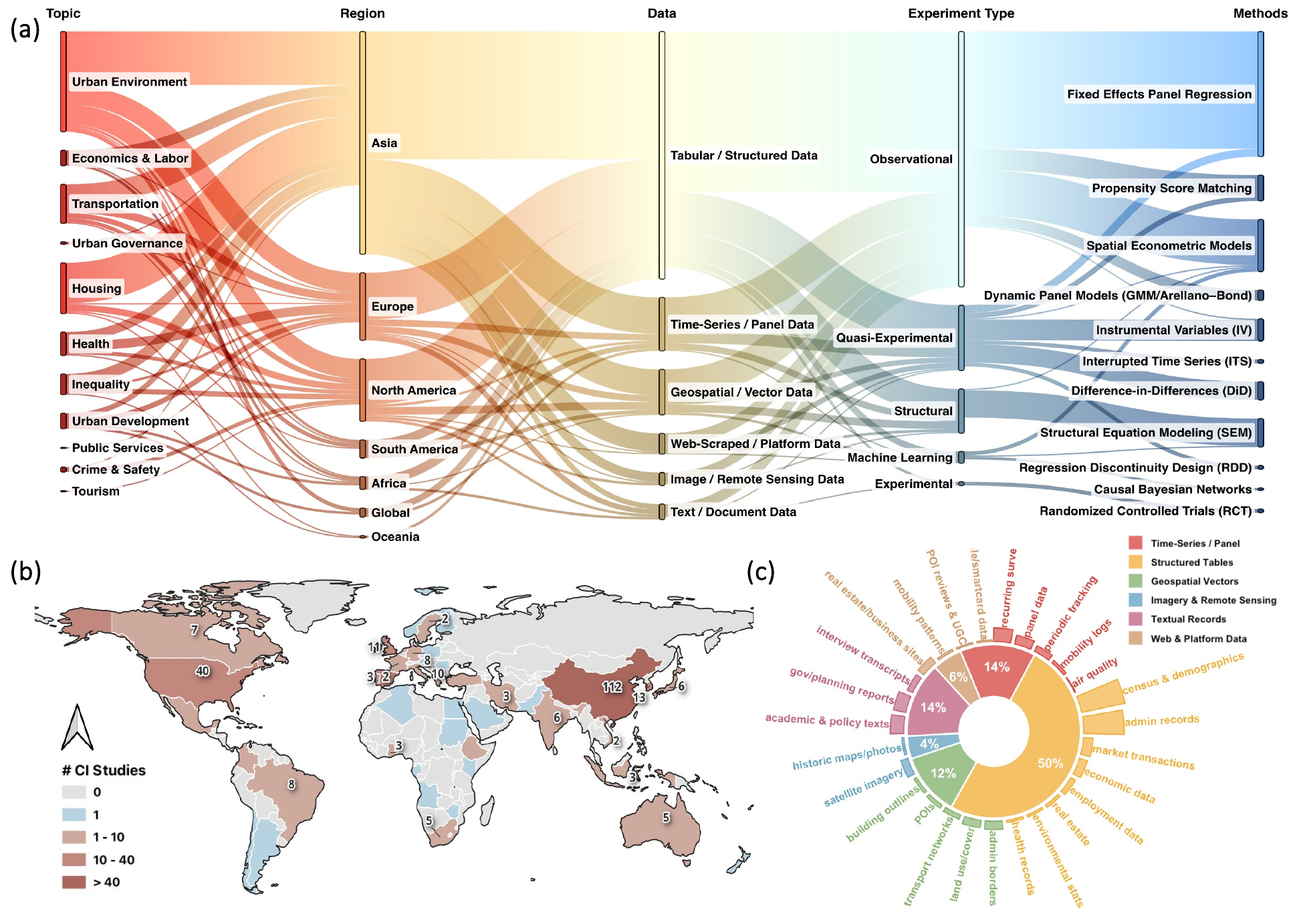}
    \vspace{-2em}
    \caption{\textbf{Key gaps in urban causal inference research.}
\textbf{(a)} Distribution of 219 studies across topics, regions, data types, experimental designs, and methods, revealing narrow topical coverage and methodological homogeneity.
\textbf{(b)} Global disparities, with over 75\% of studies from China, the United States, and Spain.
\textbf{(c)} Limited use of multimodal data, with 64\% of studies relying solely on structured tabular sources.}
    \label{fig:review}
    \vspace{-1em}
\end{figure*}

\noindent \textbf{\textit{- Analysis Procedure.}} For each paper, we extracted metadata along five dimensions: Urban Topic (e.g., housing, transportation), Geographic Focus (i.e., continent and country), Specific Data Used (e.g., panel data, satellite imagery), Causal Methodology (i.e., method family and technique), and Reproducibility (i.e., availability of code and data). The results from the analysis are presented through visualizations in Figure \ref{fig:review} (i.e., distributions of the investigated dimensions). 

\subsection{What We Found? }
\textbf{\textit{- Key trends and empirical patterns.}} Our systematic review reveals a field that is gaining momentum but remains uneven and constrained in several dimensions. While it has seen \textit{growing adoption} over the past decade, it is still marked by \textit{geographical imbalances}, \textit{a heavy reliance on structured data}, \textit{limited methodological diversity}, and \textit{poor reproducibility}, detailed as follows.

\begin{itemize}[itemsep=0pt, topsep=0pt,leftmargin=*]
    \item \textbf{A Growing Emphasis on Causal Research.}
Causal inference has attracted increasing attention from urban science scholars over the past decade. As shown in Figure~\ref{fig:trend}, in 2012, only 2 out of 67 studies (3.0\%) in \textit{Cities} used causal inference methods. By 2020, this proportion rose to 64 out of 421 studies (15.2\%). From the growing trend shown in the figure, we can see a clear shift toward evidence-based evaluation in urban policy research.

\item \textbf{Geographical Imbalances.}
As indicated in Figure~\ref{fig:review}b, urban causal research is heavily concentrated in a few countries. The top three, China (112 studies), United States (40), and Spain (16), together account for over 75\% of the total causal inference corpus. In contrast, entire regions such as Sub-Saharan Africa, South Asia, and Latin America remain significantly underrepresented. Even countries undergoing major urban transitions, such as India, Brazil, and South Africa, each have fewer than 10 studies. These imbalances raise concerns about the generalizability of findings and underscore the need for scalable solutions that lower entry barriers for underrepresented regions.

\item \textbf{Overreliance on Structured Data.}
The majority of studies rely on structured tabular datasets, including census records, economic indicators, and administrative databases. Based on our classification in Figure~\ref{fig:review}c, 64\% of studies use structured data and time series panel data exclusively. Only 12\% incorporate geospatial or vector data, 4\% utilize image or remote sensing data, and 14\% analyze textual or document-based data. This underutilization of multimodal data reflects analytical limitations, an area where AI systems can offer significant advances.

\item \textbf{Limited Methodological Diversity.}
As shown in Figure~\ref{fig:review}a, methodological approaches are dominated by conventional econometric techniques. Among the 219 papers, 38.4\% use Fixed Effects Panel Regression, 23.7\% apply Difference-in-Differences (DiD), 17.4\% use Instrumental Variables (IV), and 11.9\% rely on Propensity Score Matching (PSM). In contrast, machine learning-based methods such as causal forests appear in fewer than 3\% of studies. Bayesian models and structural equation modeling are similarly rare, despite their strengths in modeling latent and complex constructs.

\item \textbf{Lack of Transparency and Reproducibility.}
Among the 219 causal inference papers reviewed, \textbf{none} provided open access to their codebase, and fewer than 15\% offered accessible data. This lack of transparency significantly hinders reproducibility and restricts opportunities for secondary analysis or AI-assisted meta-learning.
    
\end{itemize}
\subsection{Insights from Empirical Landscape}
The findings from our systematic review point to several critical directions for designing future AI-augmented systems for urban causal research.
\textit{First}, broader topical coverage is needed. Current studies are concentrated in a narrow set of urban domains, leaving governance, inequality, and service delivery underexplored.
\textit{Second}, systems should natively support diverse and unstructured data formats, such as geospatial vectors, imagery, and textual records, which are currently underutilized but essential for capturing complex urban phenomena.
\textit{Third}, the heavy reliance on conventional econometric models suggests a need for more flexible methodological toolkits that can adapt to data properties and research questions.
\textit{Fourth}, the lack of transparency and open access underscores the importance of embedding reproducibility and accountability into system design (e.g., generating logs, versioned outputs, and shareable code).
\textit{Finally}, geographic imbalances highlight the need for systems that lower technical entry barriers, enabling participation from data-scarce or underrepresented regions.
These implications provide foundational guidance for future tools that seek to make urban causal inference more inclusive, transparent, and methodologically robust.

\section{\AutoUrbanCI framework}\label{sec:framework}
This section proposes \AutoUrbanCI, a modular and flexible framework for urban causal inference. Specifically, we begin by identifying key barreiers of current manual workflows (Figure~\ref{fig:auto-urban-ci}, \textit{left}) and then highlight the capabilities of LLMs/MLLMs that make automation feasible (Figure~\ref{fig:auto-urban-ci}, \textit{right}). Building on these two foundations, we introduce a modular, agent-based framework that structures the causal inference process into four components (Figure~\ref{fig:auto-urban-ci}, \textit{center}), each supported by different operational modes and LLM capabilities. We then present a concrete multi-agent use case to illustrate the system in action. Finally, we discuss key challenges and outline an evaluation strategy for \AutoUrbanCI.

\subsection{Motivation}
\begin{figure}[t]
    \centering
    \includegraphics[width=\linewidth]{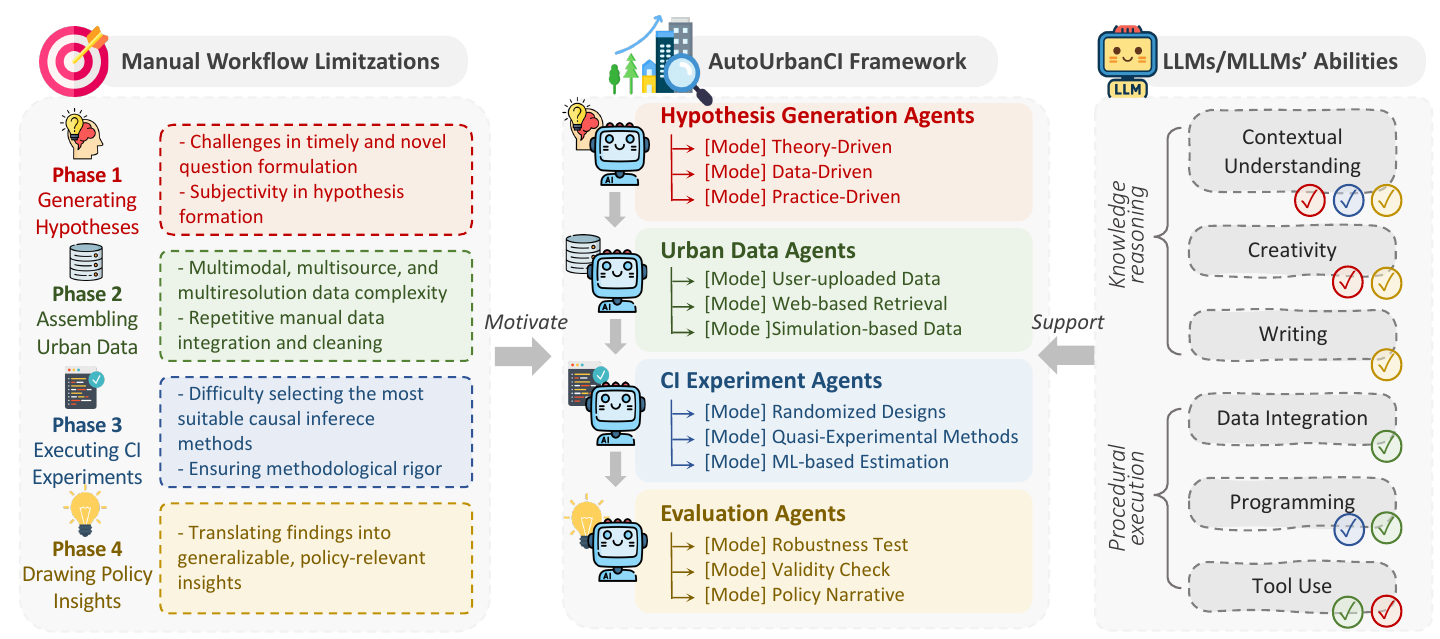}
    \vspace{-1em}
    \caption{\textbf{From limitations to design: motivating the \AutoUrbanCI framework.}. The challenges of manual urban causal workflows (\textit{left}), the modular design of \AutoUrbanCI offering agents with multiple operational modes for each stage (\textit{middle}), and the capabilities of LLMs/MLLMs that can support distinct stages, denoted by colored checkmarks (\textit{right}).}
    \label{fig:auto-urban-ci}
    \vspace{-1em}
\end{figure}

\paragraph{Why automation is needed?}
Urban causal inference requires coordination across multiple stages, from defining questions to delivering policy-relevant insights. While our systematic review in Section~\ref{sec:review} reveals increasing interest in causal methods, it also uncovers persistent limitations in current practice, i.e., overreliance on structured data, limited methodological diversity, and a lack of reproducibility. These issues stem in part from how each stage of the causal workflow is currently executed (Figure~\ref{fig:auto-urban-ci}, \textit{left}). 
In \textit{Phase 1}, generating hypotheses often relies on subjective intuition or delayed academic cycles, making it difficult to identify timely and novel research questions. 
In \textit{Phase 2}, assembling urban data is frequently repetitive and labor-intensive, requiring manual retrieval, cleaning, and integration of multi-scource (e.g., traffic sensors, social media, and environmental data) multimodal (e.g., image, spatial time series, and text), multiresolution (e.g., different smallest units or time intervals) datasets~\cite{zou2025deep,zhang2024towards}. These processes are time-consuming and difficult to generalize across cities.
\textit{Phase 3} focuses on executing causal inference experiments. Here, researchers must not only select the most appropriate estimation strategy (e.g., PSM, DiD, or IV) but also ensure alignment with the assumptions and structure of the available data, which often demands substantial methodological expertise. Finally, in \textit{Phase 4}, drawing policy insights involves synthesizing findings into accessible narratives for decision-makers, often requiring interdisciplinary translation that is rarely systematized.
These limitations motivate the need for a unified, modular framework that supports automation across all four phases while maintaining scientific rigor, flexibility, and interpretability.

\paragraph{What LLMs make possible?}
Recent advances in large language and multimodal models (LLMs and MLLMs) have introduced new possibilities for automating causal inference workflows in urban research. We organize their relevant capabilities into two core categories: \textit{knowledge reasoning} and \textit{procedural execution}, which correspond closely to the intellectual and operational demands of the \AutoUrbanCI framework.

\textit{Knowledge reasoning} encompasses the ability to interpret contextual information, generate novel ideas, and communicate structured findings. This includes semantic understanding of spatial, temporal, and causal concepts~\cite{zhao2024enhancing, zhang2024towards}, creative hypothesis formation from literature, data, or policy narratives~\cite{si2024can, chakrabarty2024art}, and the capacity to articulate experimental insights into interpretable outputs such as research summaries or policy recommendations~\cite{chia2024instructeval, baek2024researchagent}.

\textit{Procedural execution} refers to the model’s capacity to interact with external tools and data ecosystems, generate executable code, and perform end-to-end causal workflows. LLMs and MLLMs can retrieve urban datasets via APIs~\cite{yang2024give}, align multimodal inputs~\cite{gupta2023visual}, and implement causal estimation methods by synthesizing and running code across statistical and machine learning libraries~\cite{qintoolllm, li2023starcoder, huang2024mlagentbench}.

These capabilities provide the foundation for the agents within \AutoUrbanCI, enabling both cognitive support (e.g., hypothesis refinement, narrative construction) and operational automation (e.g., data acquisition, model execution) across the entire urban causal inference pipeline, as illustrated in the right portion of Figure~\ref{fig:auto-urban-ci}.

\subsection{Framework structure and agent modes}
Motivated by the above-mentioned limitations and emerging LLM/MLLMs' capabilities, we propose \AutoUrbanCI—a modular framework that structures urban causal inference into functional phases, with LLM agents assigned scoped roles and distinct modes of interaction. 
As shown in Figure~\ref{fig:auto-urban-ci} (\textit{middle}), \AutoUrbanCI is a flexible framework composed of four functional phases. Note that each phase can be supported by one or more agents, and agents can also operate across multiple phases as needed. Instead of a fixed pipeline, the framework provides a modular structure that adapts to varying data availability, task complexity, and automation needs. This flexibility allows \AutoUrbanCI to support diverse use cases, including real-time policy decisions and data-scarce environments. The agent setup described here is one example; a concrete multi-agent instantiation is introduced in Section~\ref{sec:example}. We then describe each phase with different mode in detail below.

\paragraph{Hypothesis Generation Agents.}
The first phase of the causal workflow involves identifying meaningful and timely causal questions (i.e., hypothesis) rooted in urban phenomena. Traditionally, this step relies heavily on domain expertise, often guided by literature reviews or retrospective interpretations of policy outcomes—an approach that can introduce cognitive bias and limit scalability. To address this, \AutoUrbanCI supports multiple operational modes, each tailored to different types of input sources for hypothesis generation.
\begin{itemize}[itemsep=0pt, topsep=2pt]
\item In the \textit{theory-driven} mode, agents assist in refining or critiquing human-provided hypotheses by referencing academic literature and established theoretical frameworks. 
\item The \textit{data-driven} mode identifies anomalies or shifts in sensor-based observations (e.g., spikes in traffic congestion or air pollution) that may point to candidate interventions or underlying causal mechanisms. 
\item The \textit{practice-driven} mode draws inspiration for causal questions from real-world information sources, including policy documents, urban news feeds, and social media narratives, grounding hypotheses in ongoing societal dynamics.
\end{itemize}
These modes can be invoked independently or in combination, depending on the research objective and the nature of available inputs.

\paragraph{Urban Data Agents.}
The second phase focuses on constructing datasets for downstream causal experiments, often from multi-source, multi-modal, and multi-resolution urban data. To accommodate varying levels of data accessibility and completeness, agents in this phase may operate under three distinct modes.
\begin{itemize}[itemsep=0pt, topsep=2pt]
    \item In the \textit{user-uploaded data} mode, agents process datasets provided directly by researchers, handling resolution alignment and basic preprocessing. This mode is suitable when users bring domain-specific or proprietary data into the workflow.
    \item In the \textit{web-based retrieval} mode, agents autonomously identify and collect relevant datasets from open data platforms, online repositories, or real-time APIs, guided by the variables required to test a given hypothesis. Examples of such sources include Google Street View (GSV)\footnote{\url{https://www.google.com/streetview}}, OpenStreetMap (OSM)\footnote{\url{https://www.openstreetmap.org}}, and official government portals such as NYC Open Data\footnote{\url{https://opendata.cityofnewyork.us/}} and Singapore Open Data\footnote{\url{https://data.gov.sg/}}.
    \item In the \textit{simulation-based data} mode, agents generate synthetic data to fill the missing data to support counterfactual reasoning, robustness testing, and early-stage experimentation when real-world data are costly, sparse, or unavailable. Recent works leverage LLMs to simulate structured urban behaviors—such as generating vehicle trajectories from textual descriptions~\cite{yang2025trajllm} or modeling human mobility using personas and routine profiles~\cite{ju2025trajllm,jiawei2024large}. These synthetic data environments can approximate real-world urban conditions, offering a flexible proxy for evaluating causal questions when actual interventions are infeasible.
\end{itemize}

\paragraph{CI Experiment Agents.}
Once the dataset has been prepared, agents in this phase carry out causal inference experiments to estimate the effects of candidate interventions. Depending on the structure of the data, the nature of the hypothesis, and the complexity of the treatment assignment, different estimation modes may be activated. In many cases, agents implement these experiments by generating executable code, typically in Python or R, tailored to the selected method, and then running the code to produce effect estimates. Commonly used libraries include \texttt{DoWhy} for end-to-end causal inference workflows~\cite{dowhy2019}, \texttt{EconML} for machine learning-based treatment effect estimation~\cite{econml2020}, and \texttt{CausalML} for uplift modeling and heterogeneous effect analysis~\cite{chen2020causalml}.
\begin{itemize}[itemsep=0pt, topsep=2pt]
    \item In the \textit{randomized designs} mode, agents apply experimental frameworks such as RCTs~\cite{hernan2020causal} or their approximations when random assignment is feasible, observed, or can be simulated.. In urban research, true randomization is rarely possible due to ethical, logistical, and political constraints. Thus, this mode is most often used in conjunction with simulation-based data generated in the previous phase. When treatment and control groups can be synthetically constructed, agents can estimate causal effects under strong identification assumptions.
    \item The \textit{quasi-experimental methods} mode is suitable when natural or institutional variation can be leveraged as a source of exogenous treatment. Agents operating in this mode identify applicable designs such as DiD, PSM, IV, or SCM~\cite{hernan2020causal}, depending on the data structure and policy context.
    \item In the \textit{ML-based estimation} mode, agents use data-driven models to estimate treatment effects in high-dimensional or heterogeneous settings. These may include causal forests~\cite{ronco2023exploring,ito2024examining}, deep neural networks for counterfactual inference~\cite{shalit2017estimating,louizos2017causal,shi2019adapting}, and their recent extensions to graph-structured data~\cite{wein2021graph,xia2023deciphering}. This mode is well suited for settings with treatment effect heterogeneity or limited structural assumptions.
\end{itemize}
By supporting diverse estimation strategies and automating their execution, this phase enables agents to conduct causal inference across varied urban scenarios, from controlled interventions to observational settings.

\paragraph{Evaluation Agents.}
Following causal estimation, agents in this final phase perform critical evaluations of the experimental results and help translate them into actionable insights. Depending on the research context, this process can involve technical validation, interpretability enhancement, and policy communication.
\begin{itemize}[itemsep=0pt, topsep=2pt]
    \item In the \textit{Robustness Test} mode, agents assess the sensitivity of causal estimates to modeling choices, alternative specifications, and sample heterogeneity. This includes placebo tests~\cite{eggers2024placebo}, falsification analyses, and subset re-estimation to verify whether observed effects persist under reasonable perturbations.
    \item The \textit{Validity Check} mode focuses on evaluating identification assumptions and threats to internal or external validity. Agents may examine covariate balance, assess pre-treatment trends (e.g., in DiD settings), or conduct refutation tests to identify potential confounders and post-treatment bias~\cite{hernan2020causal}.
    \item In the \textit{Policy Narrative} mode, agents assist in constructing interpretable summaries and decision-relevant narratives grounded in experimental findings. This includes generating textual explanations, translating results into human-readable formats, and mapping causal insights to policy goals and stakeholder concerns.
\end{itemize}
Together, these modes ensure that causal insights are not only statistically sound but also communicable and useful for real-world urban decision-making. Notably, we restate the \textbf{flexibility of agent coordination} across the framework. Each phase can be supported by one or more agents, and agents may operate across multiple phases as needed. The workflow is \textbf{not strictly linear}: for instance, an experiment agent may determine that an IV is required, prompting the data agent to identify a suitable instrument; or, if a PSM method is selected, the evaluation agent may test for parallel trends to confirm its validity. This modular and adaptive architecture enables agents to coordinate iteratively, refining the pipeline based on evolving task demands. We illustrate this flexibility in the following use case (Section~\ref{sec:example}).

\subsection{Multi-agent system example}\label{sec:example}

\begin{figure*}[t]
    \centering
    \includegraphics[width=1\textwidth]{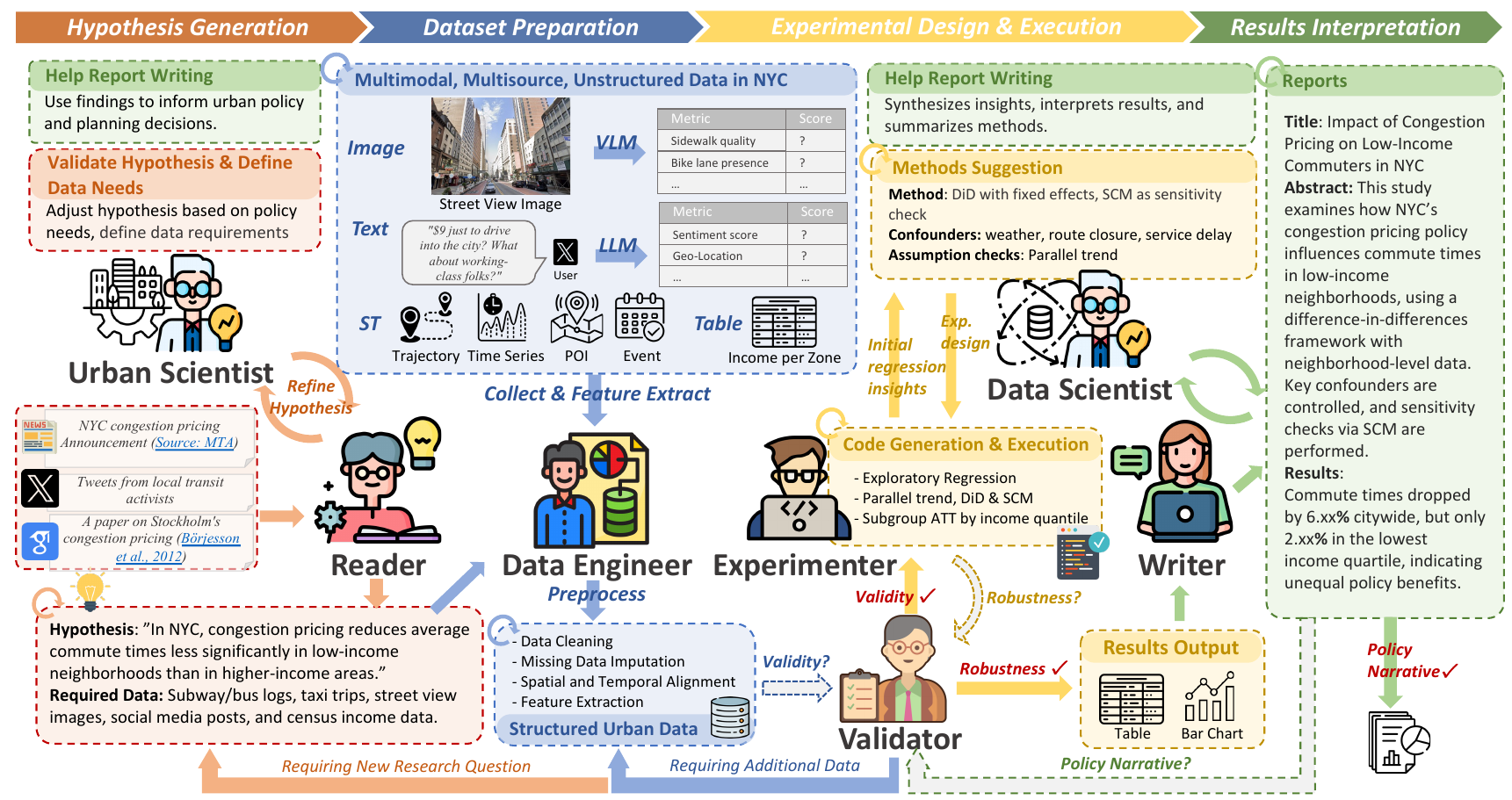}
\caption{\textbf{An example of the \AutoUrbanCI system in practice.}
    \redshaded{\textit{\textbf{Phase 1} Hypothesis Generation}}: The \AgentName{Reader} gathers policy news, social media posts, and academic papers, proposes a preliminary hypothesis, and refines it with the \AgentName{Urban Scientist} while identifying the required data types.
    \blueshaded{\textit{\textbf{Phase 2} Dataset Preparation}}: The \AgentName{Data Engineer} collects and preprocesses multimodal urban data. The \AgentName{Validator} checks the validity of the resulting dataset.
    \yellowshaded{\textit{\textbf{Phase 3} Experimental Design \& Execution}}: The \AgentName{Experimenter} runs exploratory regressions and conducts experiments under the design proposed by the \AgentName{Data Scientist}. The \AgentName{Validator} checks the robustness of the results.
    \greenshaded{\textit{\textbf{Phase 4} Results Interpretation}}: The \AgentName{Writer} synthesizes results into a structured report in collaboration with the \AgentName{Data Scientist} and \AgentName{Urban Scientist}. The \AgentName{Validator} checks the narrative alignment.
    }
    \vspace{-1em}
    \label{fig:example-case}
\end{figure*}

This section presents a concrete instantiation of \AutoUrbanCI through a multi-agent system (MAS) example, illustrating how agents collaborate across the causal workflow in a realistic research scenario. We focus on a case study investigating the impact of congestion pricing on low-income commuters in New York City (NYC), as shown in Figure~\ref{fig:example-case}. This example demonstrates one possible path through the framework:
\textit{Practice-Driven + Theory-Driven} $\rightarrow$ \textit{Web-Based Retrieval} $\rightarrow$ \textit{Quasi-Experimental Methods} $\rightarrow$ \textit{Validity Check + Policy Narrative + Robustness Test}. 
The system includes seven specialized agents—\AgentName{Urban Scientist}, \AgentName{Reader}, \AgentName{Data Engineer}, \AgentName{Validator}, \AgentName{Data Scientist}, \AgentName{Experimenter}, and \AgentName{Writer}—each responsible for a distinct function in the pipeline. Detailed roles and responsibilities are described in Appendix~\ref{app:role_agent}.

\paragraph{Phase 1: Hypothesis Generation.}
This phase includes a continuous, two-stage process to identify emerging urban challenges and translate them into researchable questions. In this example, the system monitors public discourse surrounding NYC's recently announced congestion pricing policy.
A \AgentName{Reader} agent continuously collects multimodal information from social media platforms (e.g., X), online news sources, and academic search engines. Examples include news headlines such as `New York to become first US city to have congestion charge''~\cite{bbc2024congestion}, and social media posts expressing concerns like ``\$9 just to drive into Manhattan? What about working-class folks?'' These inputs are combined with academic literature retrieved from sources such as Google Scholar on prior congestion pricing implementations in cities like Stockholm~\cite{chronopoulos2012congestion,borjesson2012stockholm}.
It proposes a preliminary hypothesis, which is then refined by the \AgentName{Urban Scientist} through contextual knowledge of NYC’s transit systems and demographic structure. The final causal hypothesis states:
\textit{``In NYC, congestion pricing reduces commute times more for higher-income commuters than for lower-income commuters, as the latter are more likely to be priced out of driving and forced to rely on public transit options.''} Relevant data sources are also identified, including subway and bus logs, taxi trips, street view images, social media posts, and census-based income records.

\paragraph{Phase 2: Dataset Preparation.}
Once the hypothesis is defined and relevant data requirements are identified, the \AgentName{Data Engineer} constructs a pipeline to collect, process, and integrate multimodal urban data into a structured, analysis-ready dataset. This involves cross-domain web-based retrieval of spatio-temporal, visual, and textual data~\cite{zou2025deep}, spanning street view imagery, transit logs (subway, bus, taxi), social media posts, and census income records~\cite{zhang2024towards}. To extract features from heterogeneous modalities, Vision-Language Models (VLMs)~\cite{bordes2024introduction} and the Segment Anything Model (SAM)~\cite{kirillov2023segment} are applied to visual inputs, while models like LLMs assist in extracting semantic attributes from text. Geo-location information is obtained either from structured metadata (e.g., geo-tags) or through named entity recognition and geocoding techniques applied to textual content~\cite{yan2024urbanclip}. Standard preprocessing steps—data cleaning, normalization, categorical encoding, and spatio-temporal alignment—ensure consistency across scales and formats.

The resulting dataset is evaluated by a \AgentName{Validator}, which checks whether it satisfies core causal requirements such as temporal coverage, treatment/control group availability, and covariate completeness~\cite{bailey2024causal}. If conditions met, it proceeds to the next phase; otherwise the system initiates an iterative refinement process, updating the data pipeline or revisiting the hypothesis as needed.

\paragraph{Phase 3: Experimental Design \& Execution.}
With a validated dataset, the \AgentName{Experimenter} first conducts a preliminary regression analysis (e.g., OLS) to provide the \AgentName{Data Scientist} with diagnostic insights on variable relevance and potential confounders. Guided by these results, the \AgentName{Data Scientist} selects appropriate causal inference methods based on the data structure and hypothesis. For the NYC congestion pricing case, it adopts a DiD approach with fixed effects as the primary strategy, complemented by a SCM as a sensitivity check~\cite{hernan2020causal}. Key confounders, e.g., weather, route closures, and service delays, are specified, and assumptions like parallel pre-treatment trends will need to be tested. The \AgentName{Experimenter} then implements the experimental design through code execution, including DiD and SCM estimation, subgroup analysis by income quantiles. The resulting outputs are reviewed by the \AgentName{Validator}, who assesses the robustness (e.g., placebo tests~\cite{eggers2024placebo}). If the results satisfy statistical and causal validity criteria, the system proceeds generation of visualizations and summary tables to the next phase; otherwise, it reverts to earlier stages for refinement.

\paragraph{Phase 4: Results Interpretation.}
The \AgentName{Writer} synthesizes outputs from the causal analysis in collaboration with the \AgentName{Urban Scientist} and \AgentName{Data Scientist}. The \AgentName{Data Scientist} outlines methodological choices and limitations, while the \AgentName{Urban Scientist} interprets effect sizes in the context of urban policy and equity. These perspectives are integrated into a structured report, combining visualizations, statistical justifications, and actionable insights to ensure policy relevance. The \AgentName{Validator} then reviews the report’s narrative coherence and policy alignment. If approved, the system proceeds to the finalized reports; otherwise, earlier stages are revisited for refinement.

\subsection{Challenges}
We outline key challenges in realizing \AutoUrbanCI as a scalable, policy-relevant system for causal analysis in urban environments.

\paragraph{Data Integration \& Quality.}
The first challenge arises from the inherent complexity of urban data, which are multi-source, multi-granular, and multi-modal. Specifically, urban data originate from highly heterogeneous sources, including social media, satellite imagery, mobility logs, environmental sensors, and policy documents. These datasets differ not only in modality (e.g., text, images, time series, and tabular formats) but also in granularity, ranging from neighborhood-level attributes to real-time street-level observations~\cite{zhang2024towards,zou2025deep}. Integrating such data presents substantial challenges in preprocessing, normalization, and semantic alignment~\cite{martinez2022data,liang2025foundation}. Recent work has explored construct urban knowledge graphs from unstructured data~\cite{liu2023urbankg,ning2024urbankgent}, which aim to represent complex urban entities and their relationships in a structured, machine-readable form. 
While promising, such efforts differ from our objective: rather than building relational graphs, our framework seeks to convert multi-modal observations into structured, temporally aligned datasets that support causal inference experiments. 

\paragraph{Maintaining Consistency in Multi-Agent Interactions.}  
MAS-based workflows often require agents to engage in multi-turn reasoning, where context from earlier stages (e.g., hypothesis generation) must persist throughout the pipeline. However, LLM-based agents frequently exhibit context drift, where key information from earlier prompts is forgotten or ignored in later stages~\cite{guo2024large}. This issue mirrors recent findings in multi-round simulations, where attention mechanisms fail to prioritize initial instructions as the context window expands~\cite{yang2024social}. Proposed solutions~\cite{touvron2023llama} offer promising directions for improving long-range coherence. Still, systematically benchmarking and improving agent-level consistency in causal inference workflows remains an open challenge.

\paragraph{Aligning Agents with Domain-Specific Causal Frameworks.}  
While \AutoUrbanCI leverages LLM agents for flexible reasoning, these models are not inherently grounded in established causal methodologies. Without explicit constraints, agents may produce plausible-sounding but methodologically invalid conclusions, particularly in quasi-experimental designs that require strict adherence to identification assumptions. Moreover, imposing rigid workflow templates may limit adaptability to diverse research designs and non-standard causal frameworks~\cite{schmidgall2025agent}. Effective alignment may involve expert-annotated prompting pipelines~\cite{wu2022promptchainer} or fine-tuning on domain-specific corpora~\cite{liang2025foundation}. Ensuring that \AutoUrbanCI agents reason within valid identification strategies, rather than relying on free-form generation, is essential for producing trustworthy and reproducible policy insights.

\subsection{Evaluation}\label{sec:evaluation}

\begin{table*}[!t]
\centering
\resizebox{\textwidth}{!}{
\begin{tabular}{>{\arraybackslash}p{0.23\textwidth}  
                >{\arraybackslash}p{0.45\textwidth} 
                >{\arraybackslash}p{0.15\textwidth}
                >{\arraybackslash}p{0.25\textwidth}
                }
                \shline
\textbf{Evaluation Type} &
  \textbf{Key Metrics} &
  \textbf{Evaluators} &
  \textbf{Representitives} \\\toprule
\multirow{3}{*}{Research Idea} &
  Novelty, Excitement, Feasibility, Effectiveness &
  Human &
  Si et al.~\citeyearpar{si2024can} \\
 &
  Significance, Clarity, Relevance, Originality, Feasibility &
  Human, LLM &
  ResearchAgent~\citeyearpar{baek2024researchagent} \\
 &
  Novelty, Value, Surprise, Relevance &
  Human &
  CoQuest~\citeyearpar{liu2024ai} \\\midrule
Method Development &
  Clarity, Validity, Rigor, Innovativeness, Generalizability &
  Human, LLM &
  ResearchAgent~\citeyearpar{baek2024researchagent}, MLR-Copilot~\citeyearpar{li2024mlr} \\\midrule
Experiment Design &
  Clarity, Validity, Robustness, Feasibility, Reproducibility &
  Human, LLM &
  ResearchAgent~\citeyearpar{baek2024researchagent} \\\midrule
Experiment Execution &
  Performance Improvement, Success Rate &
  Prototype Code &
  MLR-Copilot~\citeyearpar{li2024mlr} \\\midrule
\multirow{2}{*}{Generated Paper} &
  Experimental Quality, Report Quality, Usefulness &
  Human, LLM &
  Agent Laboratory~\citeyearpar{schmidgall2025agent} \\
 &
  NeurIPS-style Criteria &
  Human, LLM &
 The AI Scientist~\citeyearpar{lu2024ai} \\\midrule
Results &
  Novelty, Impact &
  Human, LLM &
  AI co-scientist~\citeyearpar{gottweis2025aicoscientist} \\\midrule
Automated Reviewer &
  Accuracy, F1 Score, AUC, FPR, FNR &
  OpenReview Dataset~\citeyearpar{ICLR2022-OpenReviewData} &
  The AI Scientist~\citeyearpar{lu2024ai} \\\midrule
\multirow{6}{*}{Overall System} &
  Control, Creativity, Meta Creativity, Cognitive Load, Trust &
  Human &
  CoQuest~\citeyearpar{liu2024ai} \\
 &
  Utility, Continuation, Satisfaction, Usability &
  Human &
  Agent Laboratory~\citeyearpar{schmidgall2025agent} \\
 &
  Cost (USD), Time (seconds), Success Rate &
  System Logs &
  Agent Laboratory~\citeyearpar{schmidgall2025agent} \\
 &
  Elo Rating &
  Human, LLM &
  AI co-scientist~\citeyearpar{gottweis2025aicoscientist} \\
 &
  System's Accuracy &
  GPQA Dataset~\citeyearpar{rein2024gpqa} &
  AI co-scientist~\citeyearpar{gottweis2025aicoscientist} \\
 &
  Safety &
  Curated Adversarial Dataset &
  AI co-scientist~\citeyearpar{gottweis2025aicoscientist}
\\\shline
\end{tabular}
}
\caption{\textbf{Overview of Evaluation Methods for Existing Autonomous LLM-Driven Research Works.} Summary of evaluation types, key metrics, evaluators, and representative examples across research-related LLM-based works and autonomous research frameworks.}
\vspace{-1em}
\label{tab:evaluation}
\end{table*}

Evaluating the expert-level capabilities of LLM-driven systems is inherently challenging~\cite{si2024can}. The evaluation of \AutoUrbanCI framework presents additional complexities, as it integrates multimodal data processing while maintaining the rigor required for causal inference experiments.
Unlike general LLM-based automated research systems, which are primarily designed for machine learning tasks or scientific discovery in natural sciences, \AutoUrbanCI introduces a novel paradigm tailored for urban causal research. This underscores the need to revisit the evaluation metrics for existing automated research systems, requiring evaluating their capacity to handle multimodal data, uphold the methodological rigor of causal inference, and ultimately generate meaningful causal insights that can inform urban policy and planning.

At present, evaluation in LLM-based research assistance or automated research encompasses a diverse range of quantitative and qualitative assessments, tailored to the specific tasks these systems perform. In Table~\ref{tab:evaluation}, we present an organized overview of evaluation methods applied in autonomous LLM-driven systems, along with representative examples from recent studies. According to the overview of existing evaluation methods, LLM-driven systems are assessed not only on their ability to generate novel and feasible research ideas but also on their capacity to develop sound methodologies, design robust experiments, and execute research tasks effectively. These evaluations involve both human experts and automated systems or LLM-based reviewers, ensuring a balance between domain-specific judgment and scalable validation. For instance, research outputs (e.g., generated papers and reports), are scrutinized based on clarity, relevance, and scientific quality, often leveraging hybrid approaches that integrate peer review mechanisms with LLM-based assessments~\cite{si2024can, baek2024researchagent, schmidgall2025agent, lu2024ai}.

Beyond evaluating individual research components, assessing the overall system performance is crucial to ensuring the reliability and ethical deployment of LLM-driven research systems. This evaluation encompasses multiple dimensions, including usability, interpretability, computational cost, and ethical considerations~\cite{liu2024ai, schmidgall2025agent}. In high-stakes decision-making contexts, particular emphasis is placed on safety to prevent potential misuse, such as dangerous research goals, dual-use risks, or misleading scientific claims~\cite{gottweis2025aicoscientist}. One effective approach to system evaluation is testing performance in structured problem settings, such as benchmark datasets\cite{ICLR2022-OpenReviewData, lu2024ai} or curated adversarial datasets\cite{gottweis2025aicoscientist}.

For \AutoUrbanCI, in addition to the existing evaluation methods, its inherent nature requires several additional evaluation focuses. We highlight three critical aspects: \textit{hypothesis generation significance}, \textit{data processing quality}, \textit{experiment design} and \textit{causal inference validity}. 
First, for idea generation evaluation, in addition to novelty and feasibility, it should also consider scientific rigor (i.e., ensuring asking appropriately clear and specific research questions adheres to established causal principles~\cite{bailey2024causal}) and interdisciplinary value (i.e., assessing whether they integrate insights from multiple domains, such as urban science, economics, and environmental studies, to provide innovative perspectives). 
Second, for data processing quality, its evaluation can focus on data completeness, consistency, preprocessing accuracy, scalability, and computational efficiency, ensuring robust, efficient, and effective multimodal data integration.
Third, experiment design requires assessing the system’s ability to select appropriate causal identification strategies (e.g., DiD, IV), justify assumptions, and choose valid baselines. One practical evaluation approach is to reproduce classic urban causal inference studies and compare whether the agent’s design choices align with or diverge from established methodological best practices.
Lastly, causal inference evaluation assesses causal structure feasibility, intervention validity, counterfactual consistency, and robustness to confounding. This can be achieved through synthetic data benchmarking, expert validation, or curated datasets derived from existing urban causal inference studies, ensuring that discovered causal insights are both statistically sound and practically relevant for urban policy applications.
Table~\ref{tab:our-evaluation} summarizes the full set of evaluation dimensions and highlights the ones that are particularly specific to the \AutoUrbanCI setting.

\begin{table}[!t]
\centering
\footnotesize
\resizebox{\textwidth}{!}{
\begin{tabular}{>{\arraybackslash}p{0.2\textwidth}  
                >{\arraybackslash}p{0.5\textwidth} 
                >{\centering\arraybackslash}p{0.05\textwidth}
                }
                \shline
\multicolumn{1}{c}{\textbf{Evaluation Focus}} &
  \multicolumn{1}{c}{\textbf{Key Aspects}} &
  \multicolumn{1}{c}{\textbf{Specific?}} \\\midrule
Hypothesis Quality             & Causal clarity, scientific rigor, cross-domain relevance                 & \ding{52} \\
Data Processing                & Completeness, temporal consistency, multimodal fusion quality            & \ding{52} \\
Causal Inference Rigor         & Intervention soundness, robustness checks, counterfactual consistency    & \ding{52} \\
Experiment Design              & Appropriate method choice, identification strategy, baseline comparisons & \ding{52} \\
Reproducibility \& Transparency & Log traceability, reproducible pipeline setup, transparent assumptions   &           \\
Ethical \& Social Considerations &
  Fairness, bias detection, value alignment, responsible intervention suggestion &
   \\
System Usability \& Efficiency  & Latency, scalability, agent coordination efficiency, task success rate   &    \\\shline      
\end{tabular}

}
\caption{\textbf{Evaluation dimensions for \AutoUrbanCI.} The final column marks criteria that are particularly specific to \AutoUrbanCI.}\label{tab:our-evaluation}
\end{table}

\section{Further Discussion}\label{sec:discussion}
\begin{wrapfigure}{R}{0.5\textwidth}
    \vspace{-0.5cm}
  \includegraphics[width=\linewidth]{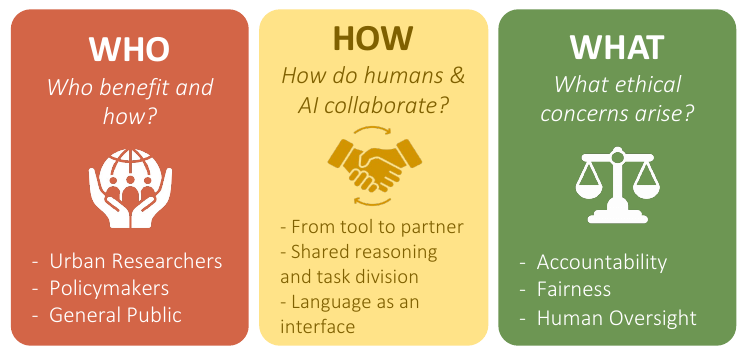}
    \caption{\textbf{Overview of key perspectives on further discussion.}. Three dimensions that shape the broader societal implications of AI-supported causal workflows in urban studies: social impact (\textit{who} benefits), human-AI collaboration paradigms (\textit{how} they interact), and ethical considerations (\textit{what} concerns arise). }
  \label{fig:discussion}
    \vspace{-0.5cm}
\end{wrapfigure}

In this section, we reflect on the broader implications of automating urban causal workflows, organized around three key perspectives: social impact, the human-AI collaboration paradigm, and ethical considerations. An overview of these dimensions is illustrated in Figure~\ref{fig:discussion}.

\subsection{Social Impact}
The proposed framework has the potential to transform scientific inquiry, policymaking, and public engagement by making causal insights more accessible, scalable, and actionable, as detailed below.

\paragraph{\textit{- Assisting Urban Researchers.}}
For urban researchers, \AutoUrbanCI serves as a powerful assistant in accelerating and refining causal analysis. This allows researchers to focus on higher-level theoretical development and policy interpretations rather than labor-intensive preprocessing and computation. Moreover, \AutoUrbanCI facilitates cross-city comparisons and large-scale causal experiments, enabling researchers to analyze urban dynamics across diverse spatial and temporal contexts. This can lead to more robust, generalizable findings that inform evidence-based urban policies and interdisciplinary research collaborations. Additionally, by standardizing causal inference workflows, \AutoUrbanCI helps ensure reproducibility and transparency in urban studies, fostering a more rigorous and scalable approach to urban causal research.

\paragraph{\textit{- Enhancing Evidence-Based Urban Policy.}}
\AutoUrbanCI provides policymakers with real-time causal analysis, allowing them to evaluate the potential impact of urban policies before implementation. By reducing reliance on correlation-driven decision-making, this automation ensures that interventions, such as traffic regulations, housing policies, or environmental protections, are based on causal insights rather than speculation. This can lead to more effective, data-driven, and proactive governance, ultimately improving urban livability, sustainability, and resource allocation.

\paragraph{\textit{- Democratizing Urban Analytics.}}
In addition, \AutoUrbanCI lowers the barriers for citizens, journalists, and grassroots organizations to explore urban issues without requiring deep technical expertise. This fosters greater civic engagement and public awareness of how cities function. Urban studies should not only seek to explain the city but also be accessible to the people who live in it —\textit{because cities can only provide something for everybody when they are created by everybody}~\cite{jacobs1961death}. This ensures that data-driven insights contribute to more inclusive and participatory urban decision-making~\cite{fox2024people}.

\subsection{New Human-AI Collaboration Paradigm}
\paragraph{\textit{- From Tool to Collaborator.}}
Traditionally, AI systems have served as passive tools in urban research workflows, used for predictive modeling~\cite{li2024urbangpt}, spatial visualization~\cite{zhang2024trafficgpt}, or data preprocessing~\cite{liang2025foundation}. However, the introduction of \AutoUrbanCI represents a shift toward a more collaborative paradigm. In this framework, AI agents are no longer limited to executing predefined tasks; instead, they actively participate in the scientific reasoning process. Rather than merely accelerating isolated stages of analysis, these agents engage in context-aware interactions that reflect an evolving role: \textit{from tool to collaborator}.
This shift invites deeper questions about how tasks and responsibilities are distributed across human–agent systems, detailed as below.

\paragraph{\textit{- Division of Reasoning and Execution.}} 
Moving beyond the traditional pipeline of human-led design, \AutoUrbanCI embraces a configurable division of labor between humans and agents. Recent systems~\cite{gottweis2025aicoscientist,li2024mlr,lu2024ai,swanson2024virtual} reveal a common design principle: human experts define goals, constraints, and evaluation criteria, while agents handle iterative generation, optimization, and evidence organization. In these systems, natural language interfaces serve as the medium for task specification, correction, and refinement, reinforcing the human’s role as an agenda-setter and critical reviewer.
In \AutoUrbanCI, this flexible structure allows agents to operate independently for routine tasks such as data preprocessing or method selection, while preserving human judgment for hypothesis formulation, identification strategy, and assumption checking. Multi-agent mechanisms such as self-critique and ranking~\cite{lu2024ai} can further enhance robustness, but the final authority remains with human users, who validate outputs and determine their scientific and policy relevance.
Rather than replacing human reasoning, this structure reinforces it: agents scaffold the causal workflow, while humans remain responsible for framing questions, defining standards, and interpreting outcomes in context.

\paragraph{\textit{- Language as Interface for Scientific Interaction.}} 
Language interfaces enable more flexible and inclusive forms of human–AI collaboration in scientific reasoning. Recent systems~\cite{zhou2022least,swanson2024virtual,gottweis2025aicoscientist} demonstrate that researchers can use natural language to specify goals, refine hypotheses, and guide iterative experimentation without relying on formal task descriptions.
This interaction model is especially valuable in urban causal research, where questions often involve loosely defined objectives and policy-relevant semantics. Through dialogic interfaces, users can engage in multi-round exchanges to clarify assumptions, compare methods, and request alternative designs. These interactions support asymmetrical collaboration, bridging technical experts, planners, and policymakers, and allow human judgment to remain embedded throughout the workflow. Moreover, by lowering technical barriers, such interfaces expand participation to civic actors such as journalists and community organizations, making urban inquiry more accessible and democratic~\cite{fox2024people}.
Language thus serves not only as an input modality, but as a reasoning interface, shaping how tasks are interpreted, decisions are justified, and causal insights are co-constructed.

\paragraph{\textit{- Toward a New Paradigm.}}
Taken together, these shifts point toward a broader rethinking of how scientific inquiry is conducted. In \AutoUrbanCI, causal analysis is no longer a static pipeline but a collaborative process, structured by agents, guided by human judgment, and mediated through language. As human and machine capabilities become increasingly entangled, urban research must move beyond automation toward shared reasoning systems that are adaptive, interpretable, and participatory. 

\subsection{Ethical Considerations}
As AI agents take on active roles across the causal workflow, from hypothesis generation to policy-facing outputs, new ethical considerations arise around responsibility and transparency, fairness and bias, and the role of human oversight.
\textit{First}, when automated inferences inform public decisions, it is essential to clarify who is accountable for the assumptions, methods, and consequences. Human-in-the-loop designs can offer oversight, but they must be structured to avoid automation bias and ensure that AI-generated outputs remain contestable. To prevent misuse of AI tools by powerful corporations or institutions for their own advantage, interpretability should enable not only technical auditing but also meaningful engagement by domain experts and affected communities.
\textit{Second}, urban datasets often encode structural inequities. If agents are trained on dominant narratives or skewed data, they may inadvertently reinforce existing biases. Systems like \AutoUrbanCI must embed mechanisms for uncertainty disclosure, sensitivity analysis, and inclusive data practices to mitigate this risk.
\textit{Third}, as AI becomes more embedded in scientific workflows, it raises broader questions around authorship, attribution, and epistemic authority. Rather than replacing human researchers, \AutoUrbanCI is designed to support more inclusive, rigorous, and reflexive forms of urban causal reasoning—where domain expertise, civic participation, and algorithmic assistance can co-exist productively.

\section{Summary \& the Road Ahead}\label{sec:sum}
\paragraph{Reimagining Urban Causal Inference.} This Perspective highlights the untapped potential of LLMs to scale urban causal research. Urban causal research plays a vital role in evidence-based policymaking, yet current workflows are often labor-intensive, fragmented, and inaccessible. Our systematic review reveals five key gaps: geographic concentration, overreliance on structured data, limited methods, poor transparency, and slow causal adoption. To address these, we introduces \AutoUrbanCI, a modular, LLM-powered framework that supports every stage of the causal pipeline, from hypothesis generation to evaluation, with an example use case. By structuring the process into discrete, automatable tasks, \AutoUrbanCI show how AI can help scale causal research while retaining scientific rigor.

\paragraph{Lowering Barriers, Broadening Participation.}
\AutoUrbanCI is more than a tool. Instead, it is a step toward a more inclusive and participatory paradigm of urban inquiry. It lowers the technical threshold for causal analysis, enabling researchers, policymakers, and civic groups to explore timely urban questions with reduced dependence on coding, manual data handling, or advanced econometrics. In doing so, it fosters greater transparency, reproducibility, and responsiveness in how cities learn and adapt.

\paragraph{The Road Ahead.}
Moving forward, we envision three key directions for future work. First, we plan to develop a deployable prototype of \AutoUrbanCI to validate its effectiveness in real-world causal research workflows. Second, we aim to refine agent capabilities, especially for handling unstructured and multimodal data and designing rigorous causal inference experiments. Third, we will explore broader deployment scenarios in collaboration with urban researchers and policymakers. We call for collective efforts across AI, urban studies, and broader quantitative social science communities to realize a more inclusive, scalable, and rigorous future for urban causal inference.

\clearpage
\bibliography{references}

\clearpage

\appendix
\onecolumn
\section{Related Works}

\subsection{Urban Causal Research}
Urban causal research aims to identify the effects of treatments, policies, and interventions on outcome variables~\cite{baum2015causal}, which is crucial for evidence-based policymaking in areas such as transportation, housing, environmental sustainability, and economic development~\cite{baum2015causal,bailey2024causal}.
Causal inference methods~\citep{hernan2020causal} thus have become indispensable tools in urban research, enabling more rigorous and data-driven policy evaluations or urban phenomenon analysis.

Randomized Controlled Trials (RCTs) are one of the commonly-used approaches for causal inference, effectively eliminating bias through random assignment. For example, ~\citet{bloom2024hybrid} employ RCTs to evaluate the impact of hybrid work-from-home policies on employee retention.
However, RCTs are often impractical in urban research due to ethical, logistical, and spatial constraints. As a result, researchers rely more on quasi-experimental methods to infer causal relationships from observational data. 
One widely used approach is Difference-in-Differences (DID), which compares outcome changes between treated and control groups over time. This method has been applied to assess the impact of high-speed rail on traffic emissions, micromobility regulations on travel behavior, and vaccine mandates on public health~\cite{lin2021impact}. 
Another key method is the Synthetic Control Method (SCM), which constructs a synthetic control unit as a weighted combination of untreated units that closely resemble the treated unit before intervention. SCM has been employed to evaluate the effects of COVID-19 lockdown policies on public sentiment and economic activity~\cite{wang2022global}.
To address endogeneity issues, Instrumental Variable (IV) methods use external instruments that influence treatment but not the outcome directly. For example, ~\citet{zheng2024impacts} have leveraged the proportion of remote-capable workers to examine the impact of remote work on vehicle miles traveled and transit ridership. In addition, Propensity Score Matching (PSM) is used to reduce selection bias by creating comparable groups based on observable characteristics, and has been applied to evaluate the economic effects of electric vehicle charging stations~\cite{zheng2024effects}. Recently, with the advancement of Machine Learning (ML), ML-based causal inference methods, such as causal forests~\cite{ronco2023exploring, ito2024examining}, have also gained popularity.
In addition, deep neural networks for counterfactual inference~\cite{shalit2017estimating,louizos2017causal,shi2019adapting}, along with their recent extensions to graph-structured data~\cite{wein2021graph}, have further expanded the toolkit for learning complex causal effects in high-dimensional settings.

However, conducting robust urban causal research requires integrating domain expertise, high-quality data, and rigorous methodologies, making it a non-trivial task.  From the data perspective, the high-volume, multi-source, and multi-modal nature of urban data demands extensive and repetitive human effort for data collection and pre-processing, reducing efficiency. 
In addition, causal methods rely heavily on subjective decisions in hypothesis formulation, variable selection, and methodological choices, which can introduce biases, overlook critical variables, and limit reproducibility.

\subsection{Multimodal Large Language Models}

Large Language Models (LLMs)~\cite{achiam2023gpt,touvron2023llama,team2023gemini,guo2025deepseek} have recently emerged as powerful tools that process and generate human-like text by leveraging vast data and transformer-based architectures~\cite{vaswani2017attention}.
Multimodal Large Language Models (MLLMs) extend the capabilities of LLMs by integrating and processing multiple data modalities, including text, images, audio, and structured data~\cite{wu2023multimodal}. 
Unlike conventional LLMs, which primarily rely on textual inputs, MLLMs leverage cross-modal reasoning to enhance understanding and decision-making in complex tasks. In other words, MLLMs can be recognized as ``universal translators'' that bridge language and sensory inputs.
Most MLLMs consist of three core components: a \textit{modality encoder} that extracts features from non-text inputs, an \textit{LLM module} that processes information, and a \textit{projector} that aligns embeddings between different modalities~\cite{fu2024mme}. These models generate responses in an autoregressive manner, incorporating both textual and non-textual cues. 

MLLMs have shown strong performance in tasks such as video captioning ~\cite{huang2024vtimellm,fu2024video} and Visual Question Answering (VQA)~\cite{guo2023images,kuang2024natural}.
In the context of urban studies, MLLMs offer the potential to analyze multimodal urban data (satellite images, traffic sensors, social media), facilitating a more holistic understanding of urban dynamics~\cite{zhang2024urban}. 
For example, UrbanCLIP~\cite{yan2024urbanclip} aligns satellite images with textual descriptors to profile a certain urban region. 
In transportation, real-time traffic data combined with social media mentions of a concert allows systems such as TrafficGPT~\cite{zhang2024trafficgpt} to diagnose congestion triggers and optimize routing. 
For disaster resilience, DisasterResponseGPT~\cite{goecks2023disasterresponsegpt} dynamically generates emergency plans by contextualizing real-time data against crisis protocols.

\subsection{AI for Scientific Discovery}

Recent advancements in AI have driven a paradigm shift in scientific discovery. 
Specifically, AI has demonstrated its potential in scientific breakthroughs, such as AlphaFold 2~\cite{jumper2021highly} in protein structure prediction, advancing drug discovery and materials science. 
Beyond specialized models, recent efforts integrate LLMs into the entire research workflow, from hypothesis generation to manuscript writing. 
Studies suggest LLM-generated feedback can be as effective as human peer reviews~\cite{liang2024can}.  
Several AI-driven systems aim to assist in different stages of scientific research: PaperQA~\cite{skarlinski2024language} aids literature search, while HypoGeniC~\cite{zhou2022least} facilitates hypothesis generation through iterative refinement and platforms like data-to-paper~\cite{ifargan2025autonomous} and Virtual Lab~\cite{swanson2024virtual} employ LLM-based agents to automate research paper writing and experimental design. 
Multi-agent systems like Coscientist~\cite{boiko2023autonomous} and The AI Scientist~\cite{lu2024ai} extend this paradigm by autonomously handling hypothesis formulation, experimental execution, and research documentation. Recently, the AI co-scientist~\cite{gottweis2025aicoscientist}, a multi-agent system built on Gemini 2.0, demonstrated its potential in drug repurposing, novel target discovery, and bacterial evolution.

Though promising, most AI-driven scientific discovery primarily focused on natural science~\cite{yan2025position}; its application to social science research remains in its early stages. 
Unlike the physical and life sciences, where AI can generate and validate hypotheses through controlled experiments, social science research involves complex, evolving human behaviors that are harder to formalize and test empirically. 
Recent efforts have begun addressing this gap. AI-assisted frameworks have been developed to uncover quantitative and symbolic models, bridging parametric and non-parametric approaches~\cite{balla2025ai}. ~\citet{manning2024automated} leverages structural causal models, where LLMs autonomously generate and test social science hypotheses through in silico experiments. Additionally, LLMs have been applied to social skill training and human-AI interaction, demonstrating potential in communication research and behavioral science~\cite{yang2024social}. More recently, LLM-driven simulations model human behavior and societal dynamics, enabling controlled studies on polarization, misinformation, and economic policies~\cite{piao2025agentsociety,piao2025emergence}. 
However, AI-driven discovery in urban causal research remains largely unexplore.

\section{Formulations}

\subsection{Causal Inference Problem}\label{app:formulation}
In a fundamental causal inference problem, the objective is to estimate the causal relationship between a set of treatment variables $T$ and an outcome $y$. For each observation $i$, the data generation process of outcome $y_i$ can be denoted as follows~\cite{baum2015causal}
\begin{equation}
    y_i = T_i \beta_i + X_i \delta_i + U_i + e_i,
\label{eq:causal_model}
\end{equation}
where $y_i$ represents the observed outcome for unit $i$, influenced by a treatment variable $T_i$, a set of observed control variables $X_i$, unobserved confounders $U_i$, and a stochastic error term $e_i$. The coefficients $\beta_i$ and $\delta_i$ capture the respective effects of the treatment and control variables on the outcome. 
In the context of urban studies, the outcome variable $y_i$ can represent an urban phenomenon, e.g., air pollution levels, traffic congestion, housing prices at a given location $i$. The treatment variable $ T_i $  typically corresponds to an urban policy intervention or infrastructure change, such as the implementation of congestion pricing, new public transit infrastructure, or zoning regulation modifications.

The primary objective is to estimate  $\beta_i$, which quantifies the causal effect of the treatment on the outcome while controlling for confounders $X_i$. Various econometric and statistical methods can be employed to estimate these coefficients, including but not limited to Ordinary Least Squares (OLS) for simple regression-based analysis, Difference-in-Differences (DiD) for policy impact evaluation, Propensity Score Matching (PSM) for balancing covariates, Instrumental Variables (IV) to address endogeneity, and Synthetic Control Methods (SCM) for counterfactual estimation.

\subsection{LLMs \& MLLMs}\label{app:llm_formulation}
Large Language Models (LLMs)~\cite{achiam2023gpt,touvron2023llama,team2023gemini,guo2025deepseek} are trained on large-scale text corpora to predict token sequences in an autoregressive manner. Given a sequence of tokens \( \mathbf{x} = (x_1, x_2, \dots, x_t) \), an LLM models the conditional probability of the next token as:
\begin{equation}
    p(x_t \mid \mathbf{x}_{<t}; \theta) = \text{softmax}(\mathbf{W} \cdot \mathbf{h}_t),
\end{equation}
where \( \mathbf{h}_t \) denotes the hidden state at time step \( t \), and \( \mathbf{W} \) is the learned weight matrix mapping hidden states to token logits. The model generates text by sampling iteratively from this distribution.

Multimodal Large Language Models (MLLMs) extend this paradigm by incorporating additional modalities, e.g., imagery~\cite{liu2023visual,yan2024urbanclip}, audio~\cite{chu2023qwen}, graph~\cite{tang2024graphgpt}, and structured charts~\cite{han2023chartllama}, alongside text. These non-text inputs are embedded into a shared representation space as \( \mathbf{e}_V \), while the textual prompt is embedded as \( \mathbf{e}_T \). The MLLM then generates output tokens \( \mathbf{y} = (y_1, y_2, \dots, y_L) \) according to:
\begin{equation}
    p(\mathbf{y} \mid \mathbf{e}_V, \mathbf{e}_T) = \prod_{t=1}^{L} P(y_t \mid \mathbf{y}_{<t}, \mathbf{e}_V, \mathbf{e}_T),
\end{equation}
where \( \mathbf{y}_{<t} \) denotes previously generated tokens.

By integrating multimodal contexts, MLLMs significantly expand the capabilities of LLMs—enabling tasks such as image captioning, visual question answering, and multimodal reasoning that are beyond the scope of text-only models.

\begin{table}[t]
    \centering
    \footnotesize
    \renewcommand{\arraystretch}{1.2}
    \begin{tabular}{c l}
        \hline
        \textbf{Symbol} & \textbf{Description} \\
        \hline
        \( \mathbf{x} = (x_1, x_2, \dots, x_t) \) & Input sequence of textual tokens \\
        \( x_t \) & Token at time step \( t \) \\
        \( \mathbf{x}_{<t} \) & Sequence of tokens preceding time step \( t \) \\
        \( \mathbf{h}_t \) & Hidden state of the LLM at step \( t \) \\
        \( \mathbf{W} \) & Weight matrix mapping hidden states to token logits \\
        \( \mathbf{e}_V \) & Embeddings from non-text modalities (e.g., images, audio) \\
        \( \mathbf{e}_T \) & Textual input embeddings \\
        \( \mathbf{y} = (y_1, y_2, \dots, y_L) \) & Generated output token sequence of length \( L \) \\
        \( p(x_t \mid \mathbf{x}_{<t}; \theta) \) & Conditional probability of token \( x_t \) given history \\
        \( P(y_t \mid \mathbf{y}_{<t}, \mathbf{e}_V, \mathbf{e}_T) \) & Conditional probability of multimodal token generation \\
        \hline
    \end{tabular}
    \caption{\textbf{Notation}. Notation used for LLMs and MLLMs formulation.}
    \label{tab:notation-llm}
\end{table}

\section{Methodology for Systematic Review} \label{app:review-method}
In Section~\ref{sec:review}, we present a comprehensive systematic review that maps the current landscape of causal inference in urban studies. The objective is to obtain a quantitative understanding of the distribution of urban causal inference studies across five key dimensions: \textit{research topic}, \textit{geographic focus}, \textit{data modality}, \textit{experiment type}, and \textit{methodological approach}. Here we detail the methodology used to conduct the review as follows. The prompts used for each step and the code implementing this procedure can be found at:
\href{https://github.com/quao627/Urban-CI-LitReview}{https://github.com/quao627/Urban-CI-LitReview}.

\paragraph{Step 1: Identify Papers That Conduct Causal Inference Studies}
Among the 2,428 papers downloaded from the \textit{Cities} journal, we identify those conducting causal inference studies using a crafted prompt with the GPT-4o model. This filtering process involves two steps. First, we exclude papers that are clearly out of scope by examining only their titles and abstracts. Second, the LLM accesses the full text of the remaining papers and labels each one accordingly. This two-step process helps reduce costs by eliminating irrelevant papers early using a low-cost method.

\paragraph{Step 2: Paper Parsing via LLMs.} 
We develop a structured prompt to guide GPT through the analysis of each paper, with the aim of extracting relevant information and supporting evidence corresponding to each dimension. To retrieve content from full-text PDFs, we utilize the OpenAI API’s file upload feature\footnote{\url{https://platform.openai.com/docs/guides/pdf-files?api-mode=chat}}, which allows direct querying over document contents. This step produced initial annotations per dimension for every paper.

\paragraph{Step 3: Iterative Label Refinement.} 
For each dimension, we design an iterative clustering pipeline that leverages GPT-4o to semantically group similar entries and propose representative labels. The algorithm identifies recurring patterns, merges equivalent concepts, and refines the terminology. This step outputs a set of standardized keywords describing each paper across the five dimensions.

\paragraph{Step 4: Human Validation.} 
To ensure accuracy and domain relevance, we manually review and refine the final set of clusters and labels using expert judgment. This final validated dataset serves as the basis for the visualizations and quantitative analyses presented in the main text.

\section{Roles and Functions of Agents in the Example Use Case}\label{app:role_agent}
This section outlines the roles and responsibilities of each agent involved in the NYC congestion pricing use case illustrated in Figure~\ref{fig:example-case}.

\begin{itemize}
    \item \AgentName{Reader}. 
    Continuously gathers multimodal information from external sources (e.g., policy announcements, social media, academic papers) to identify emerging issues and formulate preliminary hypotheses. 

    \item \AgentName{Urban Scientist}. 
    Refines hypotheses by incorporating domain expertise (e.g., local infrastructure, demographics, governance context) and evaluates their feasibility and policy relevance to ensure that the research question is well-scoped and actionable. It also suggests relevant data variables required for investigation.

    \item \AgentName{Data Engineer}. 
    Builds the data pipeline to collect, process, and align multimodal urban data (e.g., visual, textual, temporal, spatial). Responsibilities include data cleaning, missing value imputation, and spatio-temporal normalization across sources.

    \item \AgentName{Validator}. 
    Oversees quality control at multiple stages: (1) verifies data readiness (e.g., treatment/control group coverage, covariate completeness), (2) assesses robustness of results (e.g., placebo tests, sensitivity checks), and (3) ensures narrative coherence and policy alignment of the final report.

    \item \AgentName{Experimenter}. 
    Implements the experimental design by executing data analysis tasks (e.g., exploratory regressions, causal estimators) and generating outputs such as subgroup estimates and summary visualizations.

    \item \AgentName{Data Scientist}. 
    Designs the causal inference strategy by selecting appropriate identification methods (e.g., quasi-experiments, structural models), defining confounders and assumptions, and guiding the implementation plan.

    \item \AgentName{Writer}. 
    Integrates analytical findings and policy interpretations into a coherent report. It composes textual explanations, curates visual summaries, and ensures the final output communicates insights that are both rigorous and decision-relevant.
\end{itemize}

\end{document}